\documentclass[runningheads]{llncs}
\usepackage[T1]{fontenc}
\usepackage{graphicx}
\usepackage{amssymb}
\usepackage{hyperref}
\usepackage{dsfont}
\usepackage{epstopdf}
\usepackage{listings}
\usepackage{amsmath}
\usepackage{multirow}
\usepackage{soul}
\usepackage{booktabs}
\usepackage{bbding}
\usepackage{makecell}
\usepackage{stfloats}
\usepackage{xcolor}
\usepackage{makecell}
\usepackage{xspace}
\usepackage{url}
\usepackage{wrapfig}
\usepackage{multirow}
\usepackage{makecell}

\usepackage[misc]{ifsym}

\usepackage{rotating}
\usepackage{graphicx}
\usepackage{amsmath}
\usepackage{booktabs} 
\usepackage{makecell}
\usepackage{multirow}
\usepackage{tikz}
\newcommand{\xmark}{%
\tikz[scale=0.23] {
    \draw[line width=0.7,line cap=round] (0,0) to [bend left=6] (1,1);
    \draw[line width=0.7,line cap=round] (0.2,0.95) to [bend right=3] (0.8,0.05);
}}
\newcommand{\cmark}{%
\tikz[scale=0.23] {
    \draw[line width=0.7,line cap=round] (0.25,0) to [bend left=10] (1,1);
    \draw[line width=0.8,line cap=round] (0,0.35) to [bend right=1] (0.23,0);
}}

\newcommand{\RealShape}[1]{\mathcal{R}^{#1}}

\author{Huihui Xu\inst{1}\textsuperscript{*} \and
Yijun Yang\inst{1}\textsuperscript{*} \and
Angelica I Aviles-Rivero\inst{3} \and
Guang Yang\inst{4} \and 
Jing Qin\inst{5} \and 
Lei Zhu\inst{1,2}\textsuperscript{(\Letter)}}
\authorrunning{X. Huihui et al.}
\institute{The Hong Kong University of Science and Technology (Guangzhou), China 
\email{leizhu@ust.hk}\\
\and
The Hong Kong University of Science and Technology, China \and 
University of Cambridge, UK
\and
Imperial College London, UK
\and
The Hong Kong Polytechnic University, China
\\
}

\begin{document}
\title{HilbertMamba: Local-Global Reciprocal Network\\ for Uterine Fibroid Segmentation \\in Ultrasound Videos}
\maketitle

\renewcommand{\thefootnote}{}
\footnotetext{\textsuperscript{$*$} Equal contributions.}

\begin{abstract}
Regular screening and early discovery of uterine fibroid are crucial for preventing potential malignant transformations and ensuring timely, life-saving interventions. To this end, we collect and annotate the first ultrasound video dataset with 100 videos for uterine fibroid segmentation (UFUV). 
We also present Local-Global Reciprocal Network (LGRNet) to efficiently and effectively propagate the long-term temporal context which is crucial to help distinguish between uninformative noisy surrounding tissues and target lesion regions.  
Specifically, the Cyclic Neighborhood Propagation (CNP) is introduced to propagate the inter-frame local temporal context in a cyclic manner. 
Moreover, to aggregate global temporal context, we first condense each frame into a set of frame bottleneck queries and devise Hilbert Selective Scan (HilbertSS) to both efficiently path connect each frame and preserve the locality bias. A distribute layer is then utilized to disseminate back the global context for reciprocal refinement. 
Extensive experiments on UFUV and three public Video Polyp Segmentation (VPS) datasets demonstrate consistent improvements compared to state-of-the-art segmentation methods, indicating the effectiveness and versatility of LGRNet. 
Code, checkpoints, and dataset are available at \url{https://github.com/bio-mlhui/LGRNet}

\end{abstract}

\section{Introduction}
Uterine fibroids are the most common benign tumors in the female genital tract, with approximately 70\% of women at risk of experiencing such diseases throughout their lifetime~\cite{uter1}. 
Consequently, regular screening and early detection of uterine fibroids are essential for initiating timely life-saving treatments. 
Since CT and MRI examinations are expensive and harmful to human bodies, ultrasound is becoming a more popular imaging modality for clinical diagnosis. 
Recently, automatic ultrasound detection and segmentation in videos have attracted much attention from the medical community~\cite{flanet,dpstt,yang2024vivim,nega_mining}.
For example, FLA-Net\cite{flanet} presents a frequency and location feature aggregation network for ultrasound video breast lesion segmentation. 
% MS-TFAL\cite{mstfal} proposes to utilize inter-frame information to mitigate noisy label problem and improve the model robustness to noisy annotation. 
UltraDet\cite{nega_mining} proposes to aggregate the negative temporal context to facilitate filtering out false positive predictions in ultrasound video breast lesion detection. 
However, automatic uterine fibroid segmentation in ultrasound videos remains unexplored. Moreover, ultrasound video segmentation is challenged by several factors including noisy temporal motions, blurry boundaries, and changing lesion size over time. 

In this paper, \textbf{1) we collect the first ultrasound video dataset for uterine fibroid segmentation (UFUV)}, which contains 100 videos with per-frame expert annotations. To handle aforementioned challenges in ultrasound segmentation, \textbf{2) we present Local-Global Reciprocal Net (LGRNet)} to efficiently and effectively aggregate the global temporal context using a set of frame bottleneck queries. In LGRNet, \textbf{3) we incorporate Cyclic Neighborhood Propagation (CNP) and Hilbert Selective Scan (HilbertSS)} which reciprocally propagate the crucial local-global temporal context through the bottleneck queries. \textbf{4) }We conduct extensive experiments to demonstrate that LGRNet can both quantitatively and qualitatively outperform state-of-the-art segmentation methods on UFUV and three publicly available Video Polyp Segmentation datasets.
\section{Method}
\begin{figure}[h]
\begin{center}
   \includegraphics[width=1.0\linewidth]{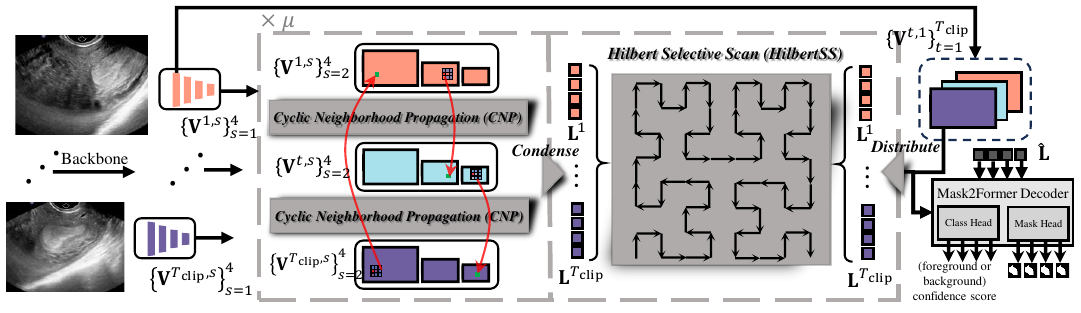}
\end{center}
\vspace{-4mm}
\caption{LGRNet architecture. We introduce the notion of \textit{frame bottleneck queries} which efficiently condense the global temporal context and distribute global context back for reciprocal local refinement.}
\vspace{-4mm}
\label{fig:framework}
\end{figure}
As shown in Fig.\ref{fig:framework}, given a video clip $\{V^t\}_{t=1}^{T_\text{clip}}$, we first devise a backbone to extract its per-frame multi-scale features $\{\{\mathbf{V}^{t,s}\in \mathcal{R}^{ H_sW_s\times c}\}_{s=1}^{S}\}_{t=1}^{T}$, where $c$ and $H_sW_s$ are dimension and resolution of the $s$-th scale. Each scale is transformed into the common dimension $c$ by a non-biased Conv2D with GroupNorm\cite{groupNorm} layer. Then, clip features of the last three scales are input to \textbf{Cyclic Neighborhood Propagation} to propagate local inter-frame motion context in a cyclic manner. $CNP$ is executed for each scale, with all scales sharing the same $CNP$ parameters. Next, for each frame, the multi-scale features are input to a $Condense$ layer and compressed into a short sequence of \textbf{frame bottleneck queries} $\mathbf{L}^t$. Bottleneck queries of all frames, i.e. $\{\mathbf{L}^t\}_{t=1}^{T_{\text{clip}}}$, are input to \textbf{Hilbert Selective Scan} to efficiently path connect all frames. Then, for each frame, global-view queries are input to a $Distribute$ layer to disseminate the global temporal context back to multi-scale features. Finally, the reciprocally encoded multi-scale features are input to a Mask2former \cite{mask2former} decoder for foreground/background classification and mask prediction. Compared with existing methods which only output single predicted mask, LGRNet can output a set of different possible mask predictions associated with a foreground lesion confidence score, which also facilities more comprehensive diagnosis.

\begin{figure}[t]
\centering
\includegraphics[width=0.95\linewidth]{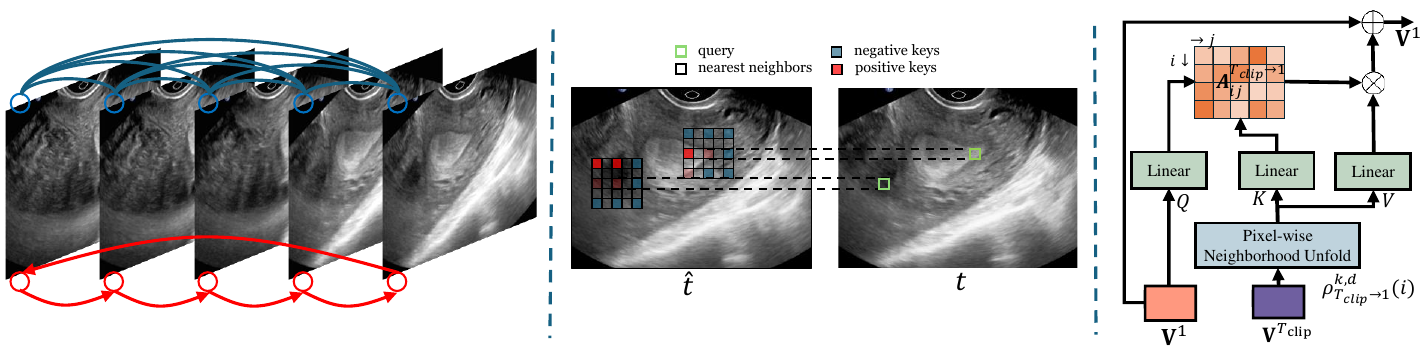}
\vspace{-4mm}
\caption{ Cyclic Neighborhood Propagation. (Left) Instead of globally token-mixing features of all frames, we enforce local cyclic inter-frame dependencies (red). (Middle) In $CNP$, each query token takes the corresponding nearest neighbors of the cyclic frame as attention keys. (Right) Detailed implementation of $CNP$.}
\vspace{-5mm}
\label{fig:cnp}
\end{figure}

\subsection{Local Cyclic Neighborhood Propagation (CNP)}
\textbf{Local CNP.} Motion priories, such as optical flow \cite{nega_mining}, can be utilized as pixel-wise guidance to propagate inter-frame temporal information. However, they incorporate additional parameters of pretrained optical flow predictor \cite{flownet} and may not generalize to ultrasound videos due to noisy and monochromatic color change. Recently, motivated by introducing locality inductive biases to vanilla attention mechanism, Neighborhood Attention (NA) \cite{hassani2023neighborhood} demonstrates that only involving nearest neighbors as attention keys can achieve comparable performance in image tasks. In this paper, we interpret inter-frame locality inductive biases as motion guidance and adapt NA to videos. We propose the Cyclic Neighborhood Propagation. $CNP$ is executed for each scale and all scales share the same set of $CNP$ parameters. We omit scale superscript $s$ in this subsection for simplicity.
As shown in Fig.\ref{fig:cnp}, for a query token $\mathbf{V}^t_i$ at $i$-th position of $t$-th frame, $CNP$ aggregates local motion information from frame $\hat{t}$ to frame $t$ as:
\begin{equation}
    CNP(\mathbf{V}^t_i, \mathbf{V}^{\hat{t}}) = \sum_{m=1}^M \mathbf{W}_m\sum_{j\in\rho_{\hat{t}\rightarrow t}^{k,d}(i)} A_{ij}^{\hat{t}\rightarrow t} \mathbf{W}_V\mathbf{V}^{\hat{t}}_{j},
\end{equation}
\begin{equation}
    \hat{t}=\begin{cases}
                t-1,& t>1\\
                T,& t=1
            \end{cases}
\end{equation}
where $M$ denotes the number of attention heads, $\rho_{\hat{t}\rightarrow t}^{k,d}(i)$ is the set of nearest neighbors w.r.t $\mathbf{V}^t_i$ at $\hat{t}$-th frame, and the nearest neighbors are defined by a kernel with size $k$ and dilation $d$. $\{A_{ij}^{\hat{t}\rightarrow t}\}_{j=1}^{|\rho_{\hat{t}\rightarrow t}^{k,d}(i)|}$ denotes the attention weights, which are the normalized dot product of query with each neighbor key, i.e. $\text{softmax}_j(
\frac{{\mathbf{V}^t_i}^T\mathbf{W}_Q^T\mathbf{W}_K\mathbf{V}_j^{\hat{t}}}{\sqrt{c}})$. Since $CNP$ is applied in each encoder layer, it enables the local inter-frame temporal information to circulate within the clip. As shown in Fig.\ref{fig:cnp}, we do not build fully connected inter-frame dependencies, since when motion changes severely and is noisy, dense connections would connect a query point to uninformative background tokens at distant frames with weak semantics, which also leads to increased computation. 

\vspace{-2mm}
\subsection{Global Hilbert Selective Scan (HilbertSS)}
\textbf{Frame Queries as Information Bottleneck.} Radiologists often need longer temporal context \cite{temporal_context} to not only decide whether a possible region is lesion or not,  but also refine their local per-frame predictions. Inspired by this behavior, we devise a set of learnable \textbf{frame bottleneck queries} $\mathbf{L}\in \mathcal{R}^{{\bar{N}\times c}}$  to first summarize each frame into a query sequence: 
\begin{equation}
    \mathbf{L}^t = Condense(\text{query}=\mathbf{L},\text{key}=\{\mathbf{V}^{t,s}\}_{s=1}^S), t=1, ..., T_\text{clip},
\label{eq:condense}
\end{equation}
where the $\textit{Condense}$ layer is implemented as a Cross Attention Layer \cite{vani_attention} with Residual connection \cite{resnet}. In the cross attention layer, the 2D multi-scale features are flattened and concatenated. Specifically, the query length is $\bar{N}$ and the key length equals to the sum of multi-scale shapes, i.e. $\sum_{s=2}^{S}H_s\times W_s$. Frame queries can be seen as bottlenecks extracting semantically rich lesion information and filtering out irrelevant and redundant features of each frame, which facilitates later efficient and effective global temporal information exchange. 

\noindent\textbf{Preliminary of Selective State Space Model.} 
\begin{figure}[t]
\centering
\includegraphics[width=0.9\linewidth]{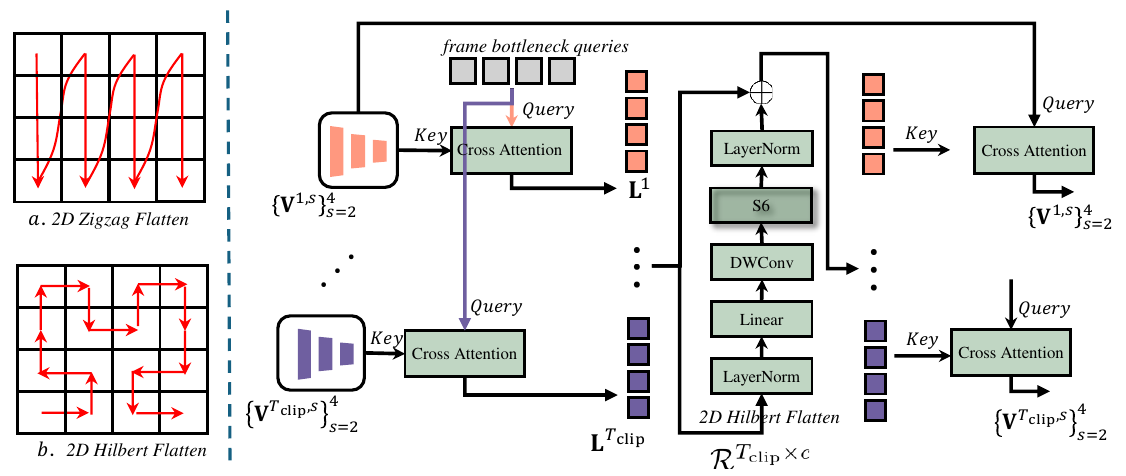}
\caption{Illustrations of $HilbertSS$. (Left) Comparison between (a) Zigzag Flatten (b) Hilbert Flatten. (Right) Detailed implementation of global part of LGRNet.}
\label{fig:hiss}
\end{figure}
\begin{figure}[t]
\centering
\includegraphics[width=0.7\linewidth]{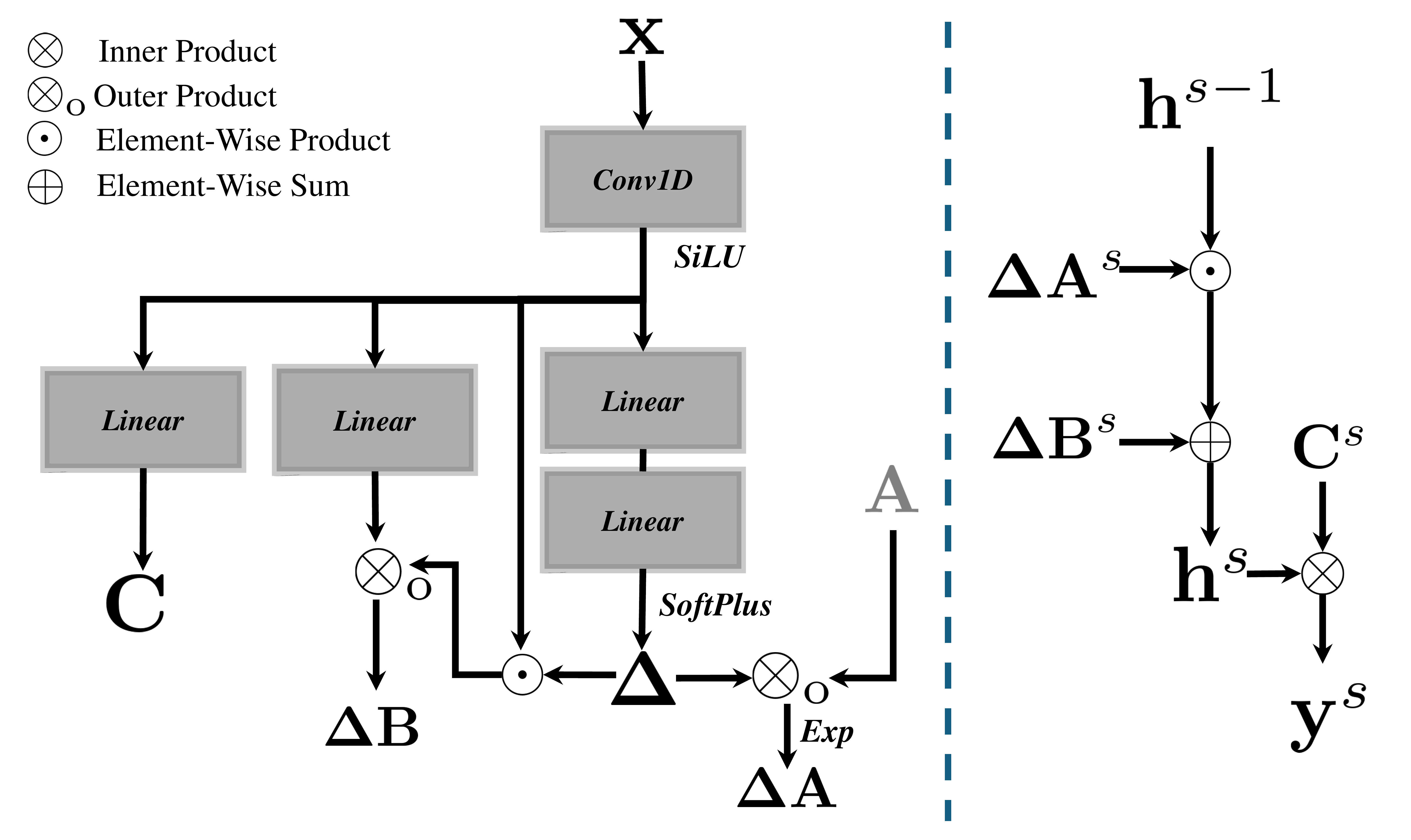}
\caption{Implementation of S6 block. (Left) Precomputation of input-dependent weights. (Right) Scanning for iterative hidden state transformation.}
\label{fig:s6}
\end{figure}
Recently, Selective State Space Model (Mamba) \cite{gu2023mamba} was proposed as a new sequence model with linear-time complexity, which even outperforms Transformers\cite{vani_attention} on some language tasks. 

Mamba is built based on the Selective Scan Block (S6). As shown in Fig.\ref{fig:s6}, each S6 block transforms the input sequence $\{ \mathbf{x}^s\in\RealShape{c}\}_{s=1}^{S}$ to $\{ \mathbf{y}^s\in\RealShape{c}\}_{s=1}^{S}$ through the precomputation stage and the scanning stage. During precomputation, a set of linear layers are used to precompute the input-dependent dynamic weights $\mathbf{\Delta B}, \mathbf{\Delta A} \in\RealShape{S\times c_{\text{state}}\times c}$, $\mathbf{C}\in\RealShape{S\times c_{\text{state}}}$, where $c_{\text{state}}$ denotes the dimension of the hidden state. The precomputation stage can be formalized as:
\begin{equation}
\begin{aligned}
    \mathbf{x} & = \text{SiLU}(\text{Conv1D}(\mathbf{x}))\\
    \mathbf{C} & = \text{Linear}_{c_{\text{state}}}(\mathbf{x})\in\mathcal{R}^{S\times c_{\text{state}}}\\
    \mathbf{\Delta} & = \text{Softplus}(\text{Linear}_{c}(\text{Linear}_{c_{\text{rank}}}(\mathbf{x}))\in\mathcal{R}^{S\times c}\\
    \mathbf{\Delta A} & = \text{exp}\{\otimes_{\text{o}}(\mathbf{\Delta}, \mathbf{A})\} \in \mathcal{R}^{S\times c_{\text{state}}\times c}\\
    \mathbf{\Delta B} & = \otimes_{\text{o}}((\mathbf{x} \odot \mathbf{\Delta}), \text{Linear}_{c_{\text{state}}}(\mathbf{x})) \in \mathcal{R}^{S\times c_{\text{state}}\times c},\\
\end{aligned}    
\end{equation}
where $\mathbf{A}$ denotes a learnable parameter of shape $c_{\text{state}}\times c$, $c_{\text{rank}}$ denotes the dimension of low-rank projection \cite{gu2023mamba}, $\text{Linear}_{*}$ denotes a linear layer transforming input to dimension $*$, $\text{SiLU}$ denotes the Sigmoid Linear Unit activation, $\text{Conv1D}$ denotes the depth-wise 1D convolution, $\otimes_{\text{o}}$ denotes the outer product, and $\text{Softplus}$ denotes the softplus activation.
To some extent, the pre-computation stage in S6 corresponds to weights computation (normalized inner product of input queries and input keys) in attention \cite{vani_attention}. Both two procedures in S6 and attention generate input-dependent, i.e. dynamic, weights for subsequent information aggregation. This characteristic enables S6 to attend to the in-context information.

During scanning, the hidden state is zero-initialized, i.e. $\mathbf{h}^0=\mathbf{0}\in \RealShape{c_{\text{state}}\times c}$. Formally, each step of hidden state transformation can be formalized as:
\begin{equation}
\begin{aligned}
    \mathbf{h}^s & = \mathbf{h}^{s-1} \odot \mathbf{\Delta A}^s \oplus \mathbf{\Delta B}^s \in \mathcal{R}^{c_{\text{state}}\times c}\\
    \mathbf{y}^s & = \otimes(\mathbf{C}^s, \mathbf{h}^s) \in \mathcal{R}^c,
\end{aligned} 
\label{eq:scan}
\end{equation}
where $\otimes$ denotes the inner product, $\oplus$ denotes the element-wise sum, $\odot$ denotes the element-wise product. Although the scanning stage is recurrent, the transition from $\mathbf{h}^{s-1}$ to $\mathbf{h}^{s}$ is associative. As shown in \cite{parallel_scan}\cite{smith2022simplified}, the scanning process can thus be parallelized to logarithmic complexity w.r.t the sequence length. 

\noindent\textbf{Global HilbertSS.} In LGRNet, we aim to token-mix the aggregated frame bottleneck queries $\{\mathbf{L}^t\}_{t=1}^{T_\text{clip}}\in \mathcal{R}^{\bar{N}\times T_{\text{clip}}\times c}$ to efficiently propagate the global clip-level temporal information. To apply S6 to 2D input, a direct approach is flattening input into a sequence using Zigzag curve as in attention-based models. However, Eq.\ref{eq:scan} implies the scanning stage is position-aware, in the sense that different scanning orders would generate different sets of hidden states. Moreover, the contextual information is propagated along the sequence only through the hidden state $\mathbf{h}^s$. Although the selection mechanism may enable boundary resetting \cite{gu2023mamba} and context filtering \cite{gu2023mamba} on some language tasks, we believe locality is more important to vision tasks, in the sense that for a local structured visual region, the hidden state should aggregate them in a group, instead of corrupting their neighboring structure in the flattened 1D sequence.

As an intuitive example shown in Fig.\ref{fig:hiss}, Zigzag scanning may corrupt the original neighborhood structure, in the sense that two tokens which are neighboring to each other on original 2D layout would be distant to each other on the flattened 1D sequence. Formally, any flattening curve can be generalized to a continuous \textit{Space Filling Curve} (SFC) $\sigma$, which maps each point $x \in$ [0, 1] to $\sigma(x)\in$ [0,1]$\times$ [0,1]. We also denote $n$ as the curve order of the Hilbert curve, which, in our discrete case, approximates to height and width of the filled grid. For any two points $x,y$ in [0,1], their \textit{Space to Linear Ratio} (SLR) is defined as:
\begin{equation}
\begin{aligned}
    \frac{|\sigma(x) - \sigma(y)|^2 }{|x - y|}.
\end{aligned}
\end{equation}
The \textit{Dilation Factor} (DF) of a SFC is defined as the upper bound of the SLR. For same two points in [0, 1], if a SFC has lower DF, their mappings will also be closer in the 2D grid, which accords with locality preserving intuition in the scanning stage. As proved in \cite{hiss_math}\cite{hilbert_3d}, the DF of Hilbert curve is 6, while the Zigzag curve is $4^n-2^{n+1} + 2$, which diverges to $\infty$ as the curve order $n$ increases. This shows that Hilbert curve could preserve the 2D locality structure, which accords with our intuition that lesions spatial-temporally close to each other should be scanned in groups and tracked together.

To execute Hilbert scan on a 2D $\bar{N}\times T_{\text{clip}}$ grid, we use the implementation of \cite{arbi} to generate the pseudo Hilbert curve. In all, the HilbertSS module can be formalized as:
\begin{equation}
    \{\mathbf{L}^t\}_{t=1}^{T_\text{clip}} = HilbertUnFlatten(S6 (HilbertFlatten(\{\mathbf{L}^t\}_{t=1}^{T_\text{clip}}))),
\end{equation}
where $HilbertFlatten$ means we use the \lstinline|torch.index_select| to flatten 2D input into a 1D sequence using the generated Hilbert indices. $HilbertUnFlatten$ means we use \lstinline|torch.Tensor.scatter_add_| to add the transformed features back to the original input according to the same Hilbert indices.

It should also be noted that existing S6-based vision models, including Vim\cite{vim}, Vmamba\cite{vmamba}, Segmamba\cite{segmamba}, and Vivim\cite{vivim}, scan the multi-scale feature maps, which include uninformative background tokens with weak semantics and thus may affect the recurrent hidden state transformation in Eq.\ref{eq:scan}. LGRNet devises a set of queries to filter out noisy information and extract the semantically rich lesion information, which would help the scanning process to effectively attend to more informative lesion regions.

\noindent\textbf{Reciprocal Local-Global Refinement.} Radiologists may use global view to refine their per-frame diagnosis. We use a $Distribute$ layer to disseminate the global temporal context back to the multi-scale features for local refinement:
\begin{equation}
    \{\mathbf{V}^{t,s}\}_{s=1}^{S} = Distribute(\text{query}=\{\mathbf{V}^{t,s}\}_{s=1}^{S}, \text{key}=\mathbf{L}^t), t=1, ..., T_\text{clip}.
\label{eq:expand}
\end{equation} 
Same to the $Condense$ layer, the $\textit{Distribute}$ layer is also implemented as a Cross Attention Layer \cite{vani_attention} with Residual connection \cite{resnet}, where the query length is $\sum_{s=2}^{S}H_s\times W_s$ and the key length is $\bar{N}$.

\vspace{-2mm}
\subsection{Decoder}
Our decoder uses the same architecture with Mask2Former\cite{mask2former}. A set of learnable \textit{temporal queries} $\hat{\mathbf{L}} \in \mathcal{R}^{\hat{N}\times c}$ are used to cross-attend to different scale features at each cross attention layer, where the query length is $\hat{N}$ and the key length is $T_{\text{clip}}\times H_s\times W_s$. The final mask is the dynamic convolution between the stride 4 scales ($s$=1) and the temporal queries. The bipartite matching loss is composed of classification (foreground/background) cross entropy loss, mask dice loss and binary cross entropy loss: $\lambda_{class} L_{class} + \lambda_{dice} L_{dice} + \lambda_{ce} L_{ce}$. During inference, the foreground classification probability can be interpreted as the lesion confidence of the corresponding predicted mask region. LGRNet can generate multiple, i.e. $\hat{N}$, mask predictions, each with a lesion confidence score. We choose the mask with the highest foreground probability as final output to compare with other methods in our experiments.

\vspace{-2mm}
\section{Experiments}
\noindent\textbf{Dataset.} We collect and annotate the first ultrasound video uterine fibroid segmentation dataset (UFUV).
Our UFUV dataset contains 100 videos and each video has 50 frames. The ultrasound videos were collected using Mindray Resona 8 and Supersonic Alxplorer. The dataset encompasses a cohort of female subjects aged between 20 to 45 years. 
We chose video that showcases at least one clearly delineated hypoechoic region (indicative of a fibroid) within the uterine wall, with a diameter exceeding 1 cm. The annotation process was rigorously conducted by two experienced gynecological ultrasound diagnosticians with over five years of professional experience. To ensure the accuracy and reliability of the annotations, the collected data underwent a cross-annotation procedure between the two diagnosticians. 
We randomly select 83 videos for training, and the remaining 17 videos are utilized for testing.

\noindent\textbf{Compared Methods.}We compare LGRNet against 9 recent state-of-the-art segmentation methods, including five image-based methods and four video-based methods.
These image-based methods are UNet++\cite{unet++}, PraNet\cite{pranet}, LDNet\cite{ldnet}, WeakPolyp\cite{weakpolyp}, and BUSSeg\cite{busseg}, while video-based methods are PNS-Net\cite{pnsnet}, DPSTT\cite{dpstt}, FLA-Net\cite{flanet}, and MS-TFAL\cite{mstfal}. 
%We implement BUSSeg and DPSTT from scratch using hyperparameters set in original paper. 
%For other methods, we inherit original hyperparameter settings in their official code for fair comparison.
%%
For each compared method, we utilize the hyperparameters settings from the original paper or their official codes for fair comparisons.

\noindent\textbf{Evaluation Metrics.} For quantitative comparison, we utilize five common metrics, including Dice Coefficient (Dice), Intersect of Union (IoU), Sensitivity, Mean Absolute Error (MAE), and S-Measure (structural similarity \cite{smeasure}). 
We also compute the inference Multiply-Accumulate Counts (MACs, GFLOPS) and the number of parameters (Params) for efficiency comparisons.

\noindent\textbf{Implementation Details.} We set $T_{clip}=6$ in our experiments. Each frame is resize to $352\times 352$. The training augmentation contains the horizontal flip, the vertical flip, and the perspective transform with magnitude 0.12. 
We use AdamW \cite{adamw} and set initial learning rate as 1e-3 with a backbone multiplier of 0.1. The multistep scheduler  the learning rate by 0.5 every 3 epochs. Gradient clipping with square norm value 1e-2 is used. We use a point sampling \cite{mask2former} with an oversampling ratio of 3.0 and importance of 0.76 for the bipartite matching mask loss computation. 
Both the number of encoder and decoder layers are set to 3. 
We use Res2Net-50\cite{res2net} as backbone, and empirically set $\lambda_{class}=2$, $\lambda_{dice}=5$, $\lambda_{ce}=2$, $c=64$, $k=5$, $d=2$, $\bar{N}=20$, $\mathcal{\mu}=3$, and $\hat{N}=10$.  
More ablation studies on hyperparameters are demonstrated in Sec.~\ref{sec:ablation}.
\begin{table*}[!t]
  \centering
  \scriptsize
  \renewcommand{\arraystretch}{0.97}
  \setlength\tabcolsep{1.2pt}
  \caption{Quantitative comparisons on our UFUV dataset.}
  \label{tab:sota_compare}
  \resizebox{0.9\textwidth}{!}{
  \begin{tabular}{ccc|ccccc|cc} 
        \hline \toprule
        Method & Publication & Type & Dice$\uparrow$ & IoU$\uparrow$ & Sensitivity $\uparrow$ & S-Measure$\uparrow$ & MAE $\downarrow$ & GFLOPs$\downarrow$ & Params$\downarrow$\\ 
        \hline
        UNet++\cite{unet++}  & TMI'19     & image       &  0.681  &   0.540 &  0.611  & 0.728 & 0.081 & 22.6 G & 29.8 M\\
        PraNet\cite{pranet}  &  MICCAI'20 & image       &  0.724  &   0.583 &  0.709  & 0.751 & 0.076 & 20.3 G & 16.1 M \\
        LDNet\cite{ldnet}    &  MICCAI'22 & image       &  0.738  &   0.588 &  0.707  & 0.753 & 0.068 & 12.6 G & 15.8 M\\
        WeakPolyp\cite{weakpolyp} & MICCAI'23  & image  &  0.725  &   0.579 &  0.682  & 0.747 & 0.075 & 5.2 G & 25.8 M\\
        BUSSeg\cite{busseg} & TMI'23  & image           &  0.740  &   0.612 &  0.711  & 0.770 & 0.066 & 23.8 G & 28.6 M\\
        \hline
        PNS-Net\cite{pnsnet} & MICCAI'21  & video       &  0.735  &   0.601 &  0.685  & 0.750 & 0.065 & 19.5 G & 15.7 M\\
        DPSTT\cite{dpstt}    & MICCAI'22  & video       &  0.738  &   0.609 &  0.707  & 0.769 & 0.065 & 24.8 G & 30.2 M\\
        FLA-Net\cite{flanet} &  MICCAI'23 & video       &  0.741  &   0.615 &  0.710  & 0.773 & 0.066 & 18.4 G & 87.6 M \\ 
        MS-TFAL\cite{mstfal} & MICCAI'23   & video      &  0.748  &   0.625 &  0.714  & 0.781 & 0.063 & 12.2 G & 24.6 M\\
      \hline
        \textbf{Ours}  & -- -- & video                                & \textbf{0.775}  & \textbf{0.658}  & \textbf{0.776} & \textbf{0.793} & \textbf{0.060} & 13.2 G & 26.6 M\\
        \hline
  \end{tabular}}
  \vspace{-4mm}
\end{table*}

\subsection{Comparisons with SOTA methods}
\begin{figure}[t]
\centering
\caption{Visual comparisons on UFUV of our network and compared SOTA methods.}
\label{fig:quality}
\includegraphics[width=0.8\linewidth]{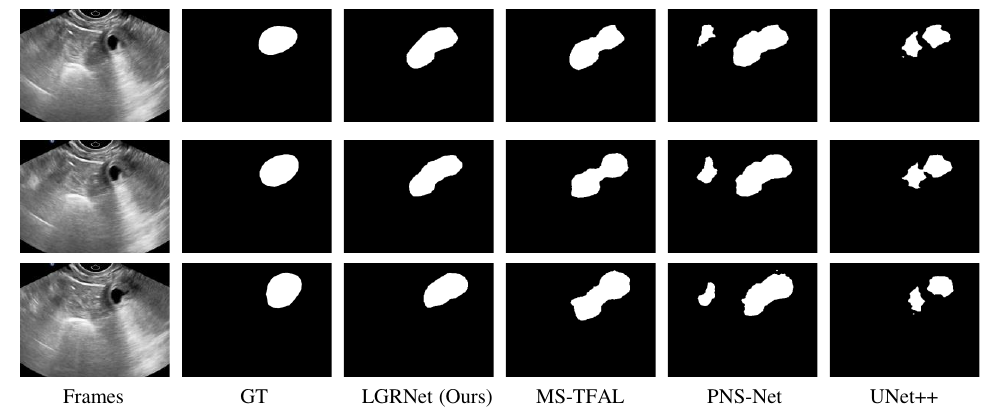}
\vspace{-4mm}
\end{figure}

\begin{table}[t!]
  \centering
  \scriptsize
  \renewcommand{\arraystretch}{0.95}
  \setlength\tabcolsep{2pt}
  \caption{Quantitative comparisons CVC-612\cite{cvc612} and CVC-300\cite{cvc300} for video polyp segmentation.
    }\label{tab:polyp1}
%   \vspace{-8pt}
\resizebox{0.9\textwidth}{!}{
  \begin{tabular}{rl||cccccc|c} 
  %funet: Spinal nerve segmentation method and dataset construction in endoscopic surgical scenarios
  \hline \toprule
  & &UNet++\cite{unet++}  & PraNet\cite{pranet} & PNS-Net \cite{pnsnet} & LDNet\cite{ldnet} & FLA-Net\cite{flanet} & MS-TFAL\cite{mstfal} & Ours\\
  & Metrics  & TMI'19 & MICCAI'20 & MICCAI'21 & MICCAI'22  & MICCAI'23 & MICCAI'23 & -- \\
\hline
 % & Speed & 108fps   & 45fps   & 20fps   & 35fps   & 97fps   & \textbf{140fps}  & \\
% \hline
\multirow{6}{*}{\begin{sideways}CVC-612-V\end{sideways}} 
& maxDice$\uparrow$     & 0.684     & 0.869   & 0.873 &   0.870   &   0.885 & 0.911 & \textbf{0.933} \\
& maxIoU$\uparrow$      & 0.570     & 0.799   & 0.800 &   0.799   &   0.814 &  0.846 & \textbf{0.877}\\
& $S_\alpha\uparrow$    & 0.805     & 0.915   & 0.923 &   0.918   &   0.920 &  0.961 & \textbf{0.947}\\
& maxSpe$\uparrow$      & 0.952     & 0.983   & 0.991 &   0.987   &   0.992 &  0.994 & \textbf{0.995}\\
& $E_\phi\uparrow$      & 0.830     & 0.936   & 0.944 &   0.941   &   0.963 &  0.971 & \textbf{0.977}\\
& $MAE\downarrow$       & 0.025     & 0.013   & 0.012 &   0.013   &   0.012 &  0.010 & \textbf{0.007}\\
\hline
\multirow{6}{*}{\begin{sideways}CVC-300-TV\end{sideways}} 
& maxDice$\uparrow$     & 0.649     & 0.739   & 0.840 &  0.835   & 0.874    & 0.891  & \textbf{0.916}  \\
& maxIoU$\uparrow$      & 0.539     & 0.645   & 0.745 &  0.741   &  0.789   & 0.810  & \textbf{0.852}\\
& $S_\alpha\uparrow$    & 0.796     & 0.833   & 0.909 &  0.898   &  0.907   & 0.912  & \textbf{0.937}\\
& maxSpe$\uparrow$      & 0.944     & 0.993   & 0.996 &  0.994   &  0.996   & 0.997  & \textbf{0.997}\\
& $E_\phi\uparrow$      & 0.831     & 0.852   & 0.921 &  0.910   &  0.969   & 0.974  & \textbf{0.986}\\
& $MAE\downarrow$       & 0.024     & 0.016   & 0.013 &  0.015   &  0.010   & 0.007  & \textbf{0.005}\\
\hline
\multirow{6}{*}{\begin{sideways}CVC-612-T\end{sideways}}  
& maxDice$\uparrow$     & 0.740     & 0.852   & 0.860 &  0.857    &  0.861   &  0.864 & \textbf{0.875}\\
& maxIoU$\uparrow$      & 0.635     & 0.786   & 0.795 &  0.791    &  0.795   &  0.796 & \textbf{0.814}\\
& $S_\alpha\uparrow$    & 0.800     & 0.886   & 0.903 &  0.892    &  0.904   &  0.906 & \textbf{0.907}\\
& maxSpe$\uparrow$      & 0.975     & 0.986   & 0.992 &  0.988    &  0.993   &  0.995 & \textbf{0.998}\\
& $E_\phi\uparrow$      & 0.817     & 0.904   & 0.903 &  0.903    &  0.904   &  0.910 & \textbf{0.915}\\
& $MAE\downarrow$       & 0.059     & 0.038   & 0.038 &  0.037    &  0.036   &  0.038 & \textbf{0.035}\\
    \bottomrule
  \end{tabular}
  }
  \vspace{-4mm}
\end{table}

\noindent
\textbf{Our UFUV dataset.} As shown in Table\ref{tab:sota_compare}, although our network does not have the smallest inference time and the smallest number of parameters, our method achieves superior performance over all compared  SOTA image/video segmentation methods on all segmentation metrics. 
Moreover, our method also achieves significant efficiency improvement compared with most methods. 
%In all, our method achieves a favorable balance between performance and speed. 
%%
Besides, Fig.\ref{fig:quality} compares the visual results produced by different methods.
Our network can more accurately segment uterine fibroid than SOTA methods, and our segmentation results are most consistent with the ground truth.
More visual comparison results are presented in the supplementary material.

%\vspace{-4mm}
%\subsection{Results on More Medical Video Object Segmentation Datasets}
\noindent
\textbf{Video Polyp Segmentation (VPS) benchmark datasets.} To further demonstrate the effectiveness of our method, we compare our network against SOTA methods on three public Video Polyp Segmentation (VPS) benchmark datasets, which are CVC-612\cite{cvc612}, CVC-300\cite{cvc300}, and SUN-SEG\cite{vps}.
For CVC-612\cite{cvc612} and CVC-300\cite{cvc300}, we follow PNS-Net\cite{pnsnet} to use the same training setting and test datasets. For SUN-SEG\cite{vps}, we follow WeakPolyp \cite{weakpolyp} to combine the "Hard (Easy) Seen" and "Hard (Easy) Unseen" split into "Hard (Easy) Testing". 
%For fair comparisons, 
%we include using PVT-v2\cite{pvtv2} as the backbone.  
As shown in Table\ref{tab:polyp1} \& \ref{tab:polyp2}, our network also achieves better metric performance than all compared methods on all three benchmark datasets, which indicates that our network has the best video polyp segmentation performance. 

\begin{table*}[!t]
  \centering
  \scriptsize
  \renewcommand{\arraystretch}{0.95}
  \setlength\tabcolsep{6pt}
  \caption{Quantitative comparisons on SUN-SEG\cite{vps} for video polyp segmentation.}
  \vspace{-2mm}
  \label{tab:polyp2}
  \resizebox{0.9\textwidth}{!}{
  \begin{tabular}{ccccccc} 
        \hline \toprule
        \multirow{2}{*}{Model} & \multirow{2}{*}{Publication} &\multirow{2}{*}{Backbone} & \multicolumn{2}{c}{Easy Testing} & \multicolumn{2}{c}{Hard Testing} \\ 
        \cmidrule(l){4-5} \cmidrule(l){6-7}
        &  & & Dice & IoU & Dice & IoU \\ 
         \hline
        PraNet~\cite{pranet}      & MICCAI'20  & Res2Net-50   & 0.689 & 0.608 & 0.660 & 0.569 \\
        2/3D~\cite{2d/3d}         & MICCAI'20  & ResNet-101   & 0.755 & 0.668 & 0.737 & 0.643 \\
        SANet~\cite{SANet}        & MICCAI'21  & Res2Net-50   & 0.693 & 0.595 & 0.640 & 0.543 \\
        PNS+~\cite{vps}           & MIR'22     & Res2Net-50   & 0.787 & 0.704 & 0.770 & 0.679 \\
        DPSTT\cite{dpstt}         & MICCAI'22  & Res2Net-50    & 0.804 & 0.725  & 0.794 &  0.709    \\
        \hline
        \multirow{2}{*}{WeakPolyp\cite{weakpolyp}} & \multirow{2}{*}{MICCAI'23} & Res2Net-50   & 0.792 & 0.715 & 0.807 & 0.727 \\
         &  & PVTv2-B2     & 0.853 & 0.781 & 0.854 & 0.777 \\ 
         \hline
        \multirow{2}{*}{FLA-Net\cite{flanet}}       & \multirow{2}{*}{MICCAI'23}  & Res2Net-50  & 0.805  &  0.723    &  0.811    &  0.730    \\
         &   & PVTv2-B2   & 0.856     &  0.784    &   0.858   &   0.781   \\
        \hline

        \multirow{2}{*}{MS-TFAL\cite{mstfal}}       & \multirow{2}{*}{MICCAI'23}  & Res2Net-50   &  0.822   &  0.742  & 0.826    &  0.751    \\
         &   & PVTv2-B2   &  0.859    &  0.792    &   0.862   &  0.788    \\
      \hline
        \textbf{Ours}  & -- -- & Res2Net-50 & \textbf{0.843}  & \textbf{0.765}  & \textbf{0.843} & \textbf{0.774} \\
        \textbf{Ours}  & -- -- & PVTv2-B2 & \textbf{0.875}  & \textbf{0.810}  & \textbf{0.876} & \textbf{0.805} \\
        \hline \toprule
  \end{tabular}}
  \vspace{-2mm}
\end{table*}

\begin{table}[!h] 
\centering
\begin{minipage}[!h]{0.46\linewidth}
    \caption{Component Analysis.} \label{tab:abl_component}
    \resizebox{\textwidth}{!}{%
    \begin{tabular}{cc|cccc}
    % \toprule
    \hline
    $CNP$ & $HilbertSS$ & Dice$\uparrow$ & IoU$\uparrow$ & S-Measure$\uparrow$ & MAE $\downarrow$\\
    \hline
    \xmark & \xmark & 0.722 & 0.581 & 0.750 & 0.074\\
    \xmark & \cmark & 0.753 & 0.633 & 0.784 & 0.062\\
    \cmark & \xmark & 0.747 & 0.626 & 0.776 & 0.067 \\
    \hline
    \cmark & \cmark & \textbf{0.775}  & \textbf{0.658}  & \textbf{0.793} & \textbf{0.060}\\
    \bottomrule
    \end{tabular}%
    }
\end{minipage}\hfill
\begin{minipage}[!h]{0.5\linewidth}
\caption{Hyperparameter Ablations} \label{tab:abl_hype}
\resizebox{\textwidth}{!}{
\begin{tabular}{c|c|cccc}
\toprule
Component & Version & Dice$\uparrow$ & IoU$\uparrow$ & S-Measure$\uparrow$ & MAE $\downarrow$\\
\hline
\multirow{4}{*}{$CNP$} & k=3, d=1 & 0.768           &  0.652          &  0.786         &  0.062\\
& k=3, d=2                        & 0.771           &  0.656          &  0.789         &  0.061 \\
& k=5, d=2                        & \textbf{0.775}  & \textbf{0.658}  & \textbf{0.793} & \textbf{0.060}\\
& k=7, d=2                        & 0.766\          & 0.647           &  0.784         & 0.064\\
\hline
 \multirow{5}{*}{$HilbertSS$} & Zigzag Scan & 0.761           &   0.639         &  0.788         & 0.060  \\
 & Hilbert Scan                             & \textbf{0.775}  & \textbf{0.658}  & \textbf{0.793} & \textbf{0.060}\\
 \cmidrule{2-6}
 & $\bar{N} = 10$  & 0.764          &   0.643         &  0.786         & 0.062 \\
 & $\bar{N} = 20$  & \textbf{0.775}  & \textbf{0.658}  & \textbf{0.793} & \textbf{0.060}\\
 & $\bar{N} = 30$  & 0.771           &   0.657         &  0.792        & 0.060  \\
\bottomrule
\end{tabular}%
}
\end{minipage}

\end{table}

\vspace{-4mm}
\subsection{Ablation Study}
\label{sec:ablation}
We conduct ablation analysis on $CNP$ and $HilbertSS$ by removing them or ablating their hyperparameters. As shown in Tab.\ref{tab:abl_component}, removing either component leads to a performance drop. Removing both components causes the model to ignore the vital temporal context information. Moreover, the global HilbertSS (0.722$\rightarrow$0.753) achieves more improvement than local $CNP$ (0.722$\rightarrow$0.747), which validates the design of frame bottleneck queries and reciprocal local-global learning. For the hyperparameter ablation, we set different kernel size $k$ and dilation $d$ for $CNP$, different selective scan strategy, and number of frame bottleneck queries $\bar{N}$ for $HilbertSS$. As shown in Fig.\ref{tab:abl_hype}, for $CNP$, both bigger kernel size and larger dilation may lead to improvement, but when much bigger kernel size is used (k=7 compared with k=5), the performance saturates. For $HilbertSS$, using Zigzag scan leads to lower performance. Moreover, using more bottleneck queries increases performance, but the performance also saturates after some threshold.  
\vspace{-2mm}
\section{Conclusion}
This paper collects and annotates the first ultrasound video uterine fibroid segmentation (UFUV) dataset, which contains 100 videos with 5,000 annotated video frames. We further propose the Local-Global Reciprocal Net (LGRNet) to efficiently aggregate global temporal context information for ultrasound video segmentation.
The $Condense$ and $Distribute$ layers with our proposed frame bottleneck queries bridge the local $CNP$ and global $HilbertSS$, and facilitate reciprocally propagating the crucial local-global temporal context information. 
Experimental results on UFUV dataset and other three public Video Polyp Segmentation (VPS) datasets show that LGRNet quantitatively and qualitatively outperforms existing state-of-the-art image and video segmentation methods.

\subsubsection{Acknowledgments} 
This work is supported by the Guangzhou-HKUST(GZ) Joint Funding Program (No. 2023A03J0671), the Guangzhou Municipal Science and Technology Project (Grant No. 2023A03J0671), and the InnoHK funding launched by Innovation and Technology Commission, Hong Kong SAR.
\subsubsection{Disclosure of Interests}
The authors declare that they have no competing interests.
\bibliographystyle{splncs04}
\bibliography{Paper-0813}
\end{document}